\documentclass{llncs}
\usepackage{amsfonts}
\usepackage{amsmath}

\begin{document}

\title{Temporal Pattern Mining from Evolving Networks}
\titlerunning{Temporal Pattern Mining from Evolving Networks}  
\author{Angelo Impedovo\inst{1} \and Corrado Loglisci\inst{1} \and Michelangelo Ceci\inst{1}}
\institute{University of Bari Aldo Moro, Department of Computer Science\\Knowledge Discovery and Data Engineering Laboratory, Bari 70125, Italy,\\
\email{\{angelo.impedovo, corrado.loglisci, michelangelo.ceci\}@uniba.it}}
\maketitle 

\section{Introduction}
Recently, evolving networks are becoming a suitable form to model
many real-world complex systems, due to their peculiarities to
represent the systems and their constituting entities, the interactions between the entities and the time-variability of their structure and properties. Designing computational models able to analyze evolving networks becomes relevant in many applications \cite{sun2012mining}.

The evolution of networks has attracted the interest of the research in Mathematics and Physics \cite{chakrabarti2006graph}, which offer theoretic mathematical models based on specific laws and global features (e.g. degree, density, diameter). However, in real-world applications, the networks often do not follow one specific model, they may reflect the combined behavior of several models and may even have exhibit unexpected characteristics. Despite the theoretic frameworks, the research of data-driven approaches is becoming promising thanks to the possibility to characterize the evolution of complex systems by means of techniques originally designed for evolving data. 
Two main categories of techniques can be recognized on the analysis of the changes, that is, \textit{clustering-based} and \textit{pattern-based}.
Clustering-based approaches \cite{berlingerio2013evolving,hayashi2015sequential} focus on the changes of network-based or node-based indicators, thus they lead to discovering changes that regards only the whole network or some nodes, without any information on the topology. 
Conversely, pattern-based approaches \cite{liu2008spotting,loglisci2015relational} rely on the \textit{frequent pattern mining} framework considering subnetworks, thus they may find valuable changes as they operate on portions of the whole network.

\section{Motivation and goals}
The goal of this research project is to evaluate the possible contribution of temporal pattern mining techniques in the analysis of evolving networks. In particular, we aim at exploiting available snapshots for the recognition of valuable and potentially useful knowledge about the temporal dynamics exhibited by the network over the time, without making any prior assumption about the underlying evolutionary schema.
\textit{Pattern-based approaches} of \textit{temporal pattern mining} can be exploited to detect and characterize changes exhibited by a network over the time, starting from observed snapshots. 

The classical task of change detection generally aims at identifying the time points in which the probability distribution of an unknown stochastic process changes. 
However, the task does not characterize the changes that have been spotted.
We aim at using pattern-based approaches in order to provide more interpretable description of such changes. 

Furthermore, the availability of training and testing data, which is a major assumption of classical machine learning algorithms, is violated when considering change point detection algorithms.
In fact, a working hypothesis is the \textit{absence of any ground truth} able to establish which data represent a change and which do not.
A quite obvious solution could be labeling, but this requires manual intervention and it is time-consuming due to the complexity of the network and to the size of the stream. 
For these reasons, unsupervised techniques are more attractive than supervised ones.

\section{Possible approaches}
Suggested approaches should: i) \textit{discover} the temporal collocation of changes, ii) \textit{characterize} the changes through the learning of descriptive models starting from observed snapshots.

The temporal collocation of changes can be found with methods based on \textit{time windows} \cite{gama2007learning}, in which network snapshots can be aggregated.
More formally, a time window of size $n$ is an ordered sequence of snapshots $W = \langle G_0, G_1, ..., G_n \rangle$ in which every snapshot $G_i$ is associated to a specific discrete time point $t_i$. 
Two time windows $W_2, W_2$ are said to be consecutive if $W_1 = \langle G_0,...,G_j \rangle$ e $W_2=\langle G_{j+1}, ..., G_k \rangle$.

An optimal partitioning of network snapshots in time windows is required. In fact, wrong partitioning could lead to miss true changes and to spot false changes. A first contribution lies in the investigation of: i) a \textit{fixed partitioning} that splits data between consecutive time windows of equal size, ii) an \textit{adaptive partitioning} that autonomously splits data in time windows delimited by time points associated to significant variations of the distributions of observed edges and nodes.

In this work time windows are used to mine frequent patterns, on their turn frequent patterns are used to denote features of the networks that are stable over time. Therefore, patterns summarize local portions of the network (subnetworks), whereas frequent patterns denote subnetworks conserved over the snapshots.
Furthermore, frequent patterns allows efficient abstraction of the network data, avoiding the need to model the changes by acting directly on the level of nodes and edges, which typically requires much more computational resources.

More formally, the relative frequency of a pattern $P$ with respect to a time window $W$ is the fraction of snapshots in $W$ of which $P$ is a subnetwork: $freq(P,W) = \frac{|\{G_i \in W | P \subseteq G_i\}|}{|W|}$. Consequently a pattern is said to be frequent in $W$ if his relative frequency in $W$ exceeds a minimum user-defined threshold value $\alpha$, $freq(P,W) > \alpha$.

If frequent subnetworks denote stable features over time, it is possible to think about the changes in terms of variations of the frequent subnetworks spotted, thus denoting changes regarding both statistical parameters and the topology of the network with certain level of statistical confidence. 
Fundamental contributions are the characterization of three types of change: i) \textit{emerging changes}, ii) \text{trend-based changes}, iii) \text{periodical changes}. Given the sequence of $m$ consecutive time windows $T = \langle W_1, W_2, ... W_m \rangle$ and a subnetwork $P$:

i) Emerging changes quantify substantial changes in the frequencies of specific subnetworks between two consecutive windows $W_i$ and $W_j$ considering the growth-rate \cite{dong1999efficient}:
\begin{equation}
	GR(P,W_i,W_j) = \frac{freq(P,W_i)}{freq(P,W_j)} \in [0, +\infty)
\end{equation}
$P$ is said to be emerging if his growth-rate exceeds a minimum user-defined threshold value $\beta$, more briefly if $GR(P,W_i,W_j) > \beta$.

ii) Trend-based changes concern gradual variations that manifest over a relatively longer sequence of windows, in which the relative frequency of conserved subnetworks monotonically increase or decrease. 
Given a sign $\psi \in \{+,-\}$ such that $\forall i \in \{1, ..., m-1\}$ then:
\begin{equation}
\begin{split}
	\psi = + \iff freq(P,W_i) < freq(P,W_{i+1}) \iff GR(P,W_{i+1}, W_i) > 1 \\
	\psi = - \iff freq(P,W_i) > freq(P,W_{i+1}) \iff GR(P,W_i, W_{i+1}) > 1
\end{split}
\end{equation}
The triple $(P,T,\psi)$ is called trend-based change. If $|T|>2$ the trend of increase/decrease is said to be global, spanning through a larger period of time, resulting more interesting. Trend discovery is a challenging task, it is unlikely that real networks exhibit subnetworks whose relative frequency monotonically vary without fluctuations over a sequence of time windows. The problem will be tackled comparing the relative frequency $freq(P,W_m)$ with an aggregate value, computed over all previous windows, such as the average relative frequency $\lambda$:
\begin{equation}
\begin{split}
	\lambda = \frac{1}{m-1}\sum\limits_{i=1}^{m-1} freq(P,W_i) \text{,} \quad \psi = \begin{cases}
		+ \iff freq(P,W_m) > \lambda\\
        - \iff freq(P,W_m) < \lambda
	\end{cases}
\end{split}
\end{equation}

iii) Periodic changes reveal regularly occurring changes over time.
The periodicity of the growth-rate is a good indicator for the relevance of a change, as this refers to events regularly occurring and then more interesting than those episodic. 
$P$ is said to be periodic of period $\pi > 0$ if emerges with the same growth-rate every $\pi$ windows:
\begin{equation}
\begin{gathered}
	GR(P,W_i,W_{i+1}) = GR(P,W_{i + k\pi},W_{i+k\pi + 1})\\
    |i - (i+k\pi)| = k\pi \text{, } k \in \mathbb{N}
\end{gathered}
\end{equation}
Periodic change discovery is a challenging task, it is unlikely that real networks exhibit specific subnetworks in a exactly periodic way. 
Main difficulties are: i) spotting exactly periodic changes of period $\pi$, ii) detecting changes that are periodically quantified by the same numerical value of growth-rate.
The limitations will be overcome by relaxing the given definition in order to characterize nearly-periodic subnetworks. Given an arbitrary tolerance level $J > 0$, the set $\Psi$ of categorical values and a mapping $\Theta : [0, +\infty) \rightarrow \Psi$, then:
\begin{equation}
\begin{gathered}
	\Theta(GR(P,W_i,W_{i+1})) = \Theta(GR(P,W_{i + k\pi},W_{i+k\pi + 1})) \\
    k\pi-J \leq |i-(i+k\pi)| \leq k\pi+J  \text{, } k \in \mathbb{N}
\end{gathered}
\end{equation}
instead of the equality test based on raw numerical values of the growth-rate, an equality test between categorical values is preferred.

\section{Final remarks and ongoing works}
Evolving networks are powerful tools used to describe the evolution of real-world complex systems, their analysis may help to better comprehend the temporal dynamics and the evolution of such complex systems.
The pattern-based representation, together with considerations about the monotonicity and periodicity of the growth-rate, may help to better characterize the relevance of the changes.
The application of temporal pattern mining approaches to the analysis of evolving networks deserves more investigation. We are currently performing experiments to evaluate the effectiveness of the emerging changes discovery as well as a distributed implementation of the proposed methods, in order to be able to process large-scale networks.

\bibliography{bibliografia.bib}

\begin{thebibliography}{1}

\bibitem{sun2012mining}
Yizhou Sun and Jiawei Han.
\newblock Mining heterogeneous information networks: principles and
  methodologies.
\newblock {\em Synthesis Lectures on Data Mining and Knowledge Discovery},
  3(2):1--159, 2012.

\bibitem{chakrabarti2006graph}
Deepayan Chakrabarti and Christos Faloutsos.
\newblock Graph mining: Laws, generators, and algorithms.
\newblock {\em ACM computing surveys (CSUR)}, 38(1):2, 2006.

\bibitem{berlingerio2013evolving}
Michele Berlingerio, Michele Coscia, Fosca Giannotti, Anna Monreale, and Dino
  Pedreschi.
\newblock Evolving networks: Eras and turning points.
\newblock {\em Intelligent Data Analysis}, 17(1):27--48, 2013.

\bibitem{hayashi2015sequential}
Yu~Hayashi and Kenji Yamanishi.
\newblock Sequential network change detection with its applications to ad
  impact relation analysis.
\newblock {\em Data Mining and Knowledge Discovery}, 29(1):137--167, 2015.

\bibitem{liu2008spotting}
Zheng Liu, Jeffrey~Xu Yu, Yiping Ke, Xuemin Lin, and Lei Chen.
\newblock Spotting significant changing subgraphs in evolving graphs.
\newblock In {\em Data Mining, 2008. ICDM'08. Eighth IEEE International
  Conference on}, pages 917--922. IEEE, 2008.

\bibitem{loglisci2015relational}
Corrado Loglisci, Michelangelo Ceci, and Donato Malerba.
\newblock Relational mining for discovering changes in evolving networks.
\newblock {\em Neurocomputing}, 150:265--288, 2015.

\bibitem{gama2007learning}
Joao Gama and Mohamed~Medhat Gaber.
\newblock Learning from data streams: Processing techniques in sensor networks.
\newblock {\em ISBN 3540736786, 9783540736783}, pages 31--33, 2007.

\bibitem{dong1999efficient}
Guozhu Dong and Jinyan Li.
\newblock Efficient mining of emerging patterns: Discovering trends and
  differences.
\newblock In {\em Proceedings of the fifth ACM SIGKDD international conference
  on Knowledge discovery and data mining}, pages 43--52. ACM, 1999.

\end{thebibliography}
\bibliographystyle{unsrt}

\end{document}